# Learning predictive models for combinations of heterogeneous proteomic data sources

**Michal Valko, Richard Pelikan and Milos Hauskrecht**
Computer Science Department
University of Pittsburgh, Pittsburgh, Pennsylvania

## ABSTRACT

*Multiple technologies that measure expression levels of protein mixtures in the human body offer a potential for detection and understanding the disease. The recent increase of these technologies prompts researchers to evaluate the individual and combined utility of data generated by the technologies. In this work, we study two data sources to measure the expression of protein mixtures in the human body: whole-sample MS profiling and multiplexed protein arrays. We investigate the individual and combined utility of these technologies by learning and testing a variety of classification models on the data from a pancreatic cancer study. We show that for the combination of these two (heterogeneous) datasets, classification models that work well on one of them individually fail on the combination of the two datasets. We study and propose a class of model fusion methods that acknowledge the differences and try to reap most of the benefits from their combination.*

## INTRODUCTION

Multiple novel technologies that measure expression levels of complex protein mixtures in the human body from a variety of specimen, such as blood or urine, offer a great potential for improved detection and understanding of the disease. With the recent increase in the number of these technologies the challenge becomes the understanding of individual and combined benefits of data generated by these technologies. As the possibilities increase, so does the data production and researchers are pressed to find the best methods suitable for their combined analysis. In this work we study two such technologies: [1] multiplexed protein arrays, and [2] MS spectrometry proteomic profiling, and analyze their individual and combined benefits in context of disease detection.

**Luminex arrays.** Conventional assay methodologies, such as ELISA, RIA, or PCR permit the analysis of a single, or a limited number of markers. Assay techniques that can simultaneously examine a large number of pre-selected proteomic biomarkers from a small amount of serum are hence of great potential value for better characterization of the disease. Luminex Corporation (http://www.luminexcorp.com) has recently introduced a novel protein array system, xMAP®, which allows for simultaneous assessment and quantitation of up to 100 soluble analytes in one sample. These analytes are carefully picked and include immunogenic and growth factor markers such as: IL-2, IL 8, MMP-7, MMP-12, IP-10 and others.

**Time-of-flight mass spectrometry (TOF-MS)** has become a widely used tool for the rapid analysis of protein-mass content in a variety of biospecimen (blood, urine, cell-lysates). Thousands of mass measurements of molecules present in the sample appear along the profile. Despite the overlap in mass signatures of these species, it is possible to find discriminatory patterns which distinguish the profiles of diseased and healthy patients. The technology has been used to successfully find patterns and predictive models for several diseases, including cancer [1, 2, 3, 4], arthritis [5], ALS [6] etc.

**Disease detection.** Classification learning methods allow us to extract models that permit detection of a disease from data. Unfortunately, a single machine learning model or a method that fits all possible datasets and consistently gives the best possible performance does not exist. We investigate the individual and combined potential of data generated by these two technologies on detecting disease in patients with pancreatic cancer by learning and testing a variety of classification models. The samples are based on the study conducted at the University of Pittsburgh Cancer Institute and data used in [7].

**Challenges.** Luminex and SELDI-TOF-MS technologies provide data that may indicate through variety of patterns the presence or absence of the disease. However, the differences in the nature of data generated by them raise concerns that they are not easy to combine into a single machine learning model. First, the immunogenic and growth factor proteomic markers measured by Luminex xMAP® technology are likely to be biologically independent of SELDI-TOF-MS profile features and patterns, thus providing complementary information about the disease. Second, SELDI-TOF-MS data are high dimensional and typically involve hundreds of measurements, while Luminex arrays are restricted to at most 100 probes (typically 30-60 probes) of pre-selected species. Third, MS-data tend to exhibit a large amount of signal correlates, while Luminex is a collection of independent probes with fewer correlations.

We show that features of the two data sources can be successfully captured by two different machine learning models. Second, we show and that a naïve combination of the data from both data sources does not lead to improvements in any of the two models and in one instance the results are much worse. To reap the benefits of the combination of the data-sources we propose and study machine learning models that acknowledge the differences in the two data streams and demonstrate their combined classification improvement.

## DATA

The pancreatic dataset used in this study consists of 106 samples, with 56 cases and 53 smoking- age and gender matched controls collected at the University of Pittsburgh Cancer Institute. There are two different data sources for these samples: 1554 peaks from SELDI-TOF-MS profiles and 30 measurements from Luminex data.

**Luminex array data.** *Luminex Corporation's* xMAP® technology uses polystyrene microspheres internally dyed with two spectrally distinct fluorophores to create a family of 100 spectrally addressed bead sets. Each bead set can be conjugated with a capture antibody specific for a unique target protein. In a multiplexed assay, antibody-conjugated beads are allowed to react with sample (plasma, serum or cell culture supernatant). After washing, secondary phycoerythrin (PE) conjugated antibodies are added to to form a capture sandwich immunoassay. The assay solution was analyzed by the array Bio-Plex reader from Bio-Rad, which obtains two readings for every single bead: one that identifies a fluorochrome labeled bead, and another that measures the amount of PE fluorescence of the PE-conjugated detection antibody associated with the bead. The amount of PE fluorescence is proportional to the amount of analyte captured in the immunoassay. Bio-Plex Manager software (Bio-Rad) correlates each bead set to the assay reagent that has been coupled to it and estimates the concentration of each analyte in the sample. Using the assay, thousands of beads can be analyzed in seconds, allowing up to 100 analytes to be measured in a 96-well microplate in one hour.

**MS proteomic profiles** Mass spectrometry proteomic profiling of serum samples was performed using SELDI-TOF-MS profiling technology (Ciphergen Biosystems, Inc.) The samples from the study were analyzed are spotted in duplicate on the IMAC-3 ProteinChip® array in blinded combined case/control batches. The protocols for the preparation and loading of serum samples for SELDI-TOF-MS analysis are specific for the ProteinChip$_r$ Arrays (Ciphergen Biosystems, Inc.). Fully automated BioMek2000$_r$ protocols for processing of IMAC3 ProteinChip$_r$ Arrays are presently being utilized in the UPCI Clinical Proteomics Facility. Protocols for automated processing of these ProteinChip$_r$ Arrays, as well as performing mass spectrometry and preprocessing of the spectral data for analysis, have been derived and optimized from protocols implemented and validated in this facility.

To eliminate systematic biases and noise in the MS data we apply five preprocessing steps implemented in the ***Proteomic data analysis package (PDAP)*** [7]: (1) variance stabilization, (2) baseline correction, (3) smoothing, (4) intensity normalization, and (5) profile alignment steps. Briefly, from many choices offered and implemented by PDAP we applied the following preprocessing choices: cube-root variance stabilization, PDAP's baseline subtraction routine based on the local moving window of width 200 time-points, Gaussian-kernel smoothing, peak-based dynamic programming alignment and the total ion current normalization restricted to the range of 1500-20000 Daltons. The qualities of profiles were tested beforehand on raw MS profile readings. None of the profiles used exhibited total ion current (TIC) that differed by more than two standard deviations from the mean TIC, which is our quality-assurance /quality-control threshold for sample exclusion.

**Peak selection.** The majority of proteomic data analyses in literature restrict their attention only to information represented in the peaks of the signal. Our ***peak selection procedure*** works with the mean profile obtained by averaging all profiles the training data. The approach is robust enough even if a specific peak is not recorded in all profiles, whilst it tends to average out random signal fluctuations. The peak locations are identified by calculating the signal derivatives. The peak intensities are average of readings in a local neighborhood of the peak location. Such a method reduces the chance of a noisy reading at a single *m/z* position.

## METHODS OF ANALYSIS

Our aim is to build a classification model $f : \mathbf{X} \rightarrow Y$ that can, with a high accuracy, assign correct class labels Y (case or control) to profiles (**X**) collected for patients in the study. We adopt a machine learning approach in which the model is learned and evaluated from the data in the study.

A number of different models and algorithms built for the classification learning task exist. The methods include: linear discriminant analysis, CART, support vector machines, logistic regression, Naïve Bayes and others. However, many of these off-shelve classification models are not immediately applicable to analysis of high dimensional proteomic datasets. The main reason is a small sample size of studies combined with the dimensionality of the data. This lead to a model overfit where every sample in the dataset is fit via one or more parameters which in turn affects model's generalization performance. To address the problem our investigations have primarily focused on two classification models with robust generalization performances on both lower and high dimensional data: Support vector machines (SVMs) and Random Forests. These methods are currently available in our PDAP package [7].

The **linear support vector machine** or SVM [8, 9] learns a linear decision boundary that separates the *n*-dimensional feature space into 2 partitions. The boundary is a hyperplane given by the equation $\mathbf{w}^T\mathbf{x} + w_0 = 0$, where $\mathbf{w}$ is the normal to the hyperplane, and $w_0$ is the distance separating the "support vectors" - representative samples from each class which are most helpful for defining the decision boundary. The parameters of the model, $\mathbf{w}$ and $w_0$, can be learned from data through quadratic optimization using a set of Lagrange parameters $\hat{\alpha}_i$. These parameters allow us to redefine the decision boundary as

$$\hat{\mathbf{w}}^T\mathbf{x} + w_0 = \sum_{i \in SV} \hat{\alpha}_i y_i (\mathbf{x}_i^T \mathbf{x}) + w_0$$

where only samples in the support vector contribute to the computation of the decision boundary. The support vector machine comes with built in regularization abilities which perform very well when high dimensional datasets needs to be analyzed. [8, 17]

**Random forest** [11] is an ensemble classifier [12] that combines the results of multiple decision trees classifiers through averaging. The decision trees in the ensemble are built using Classification and Regression Tree (CART) [13] or other tree-building algorithms, but the tree building is enriched through random feature selection process. Decision trees tend to work well for lower dimensional data and their advantage is the ability to capture various differences in between the groups via multiple non-linear decision splits. Random feature selection lets us to consider subsets of features rather than all features when learning the decision classifier which is believed to be an advantage if the dimensionality of the data is large when compared to the number of samples.

## Combination of data sources

The simplest method to combine data from multiple data sources for classification purposes is to merge them together. A classifier is build for all data at once. The advantage of the approach is that no new models need to be considered, the limitation is that the performances of classifiers one can achieve on data corresponding to individual data sources may become hard to improve.

There is no universal classifier that works the best for all possible data sources. [16] Instead one often sees one classifier dominating the other on certain type of data. [16] Hence it is possible that different (base) classifiers exhibit the dominance for the data sources we have. The key challenge is to develop classification models that let us fit individual data sources well and after that combine their performance to improve the overall classification accuracy.

To address this problem we investigate two methods for combining different classification models: (1) model inclusion and (2) model composition. Both of these models take advantage of an opportunity to build soft discriminative projections from existing classification models. For SVM, it is a distance from a hyperplane ($\mathbf{w}^T\mathbf{x} + w_0 = 0$). Likewise, in RF, we can compute the ratio of the trees that favor a particular class. We use these soft discriminative functions to design a compound classifier for the combined data sources.

**Model inclusion:** First, we can use a soft output from one classifier as an extra feature for another. For example: We take SELDI data and learn a SVM for it. Then we the (train) data projected on a hyperplane and include this (1 dimensional) vector as an extra feature with the Luminex data. Eventually, we learn a RF classifier for the merged data. Conversely we can take tree ratio from the Luminex data and add it as and extra feature with SELDI for SVM.

**Model composition** approach builds a two layer classification model. On the first level a base classifier for a given data type is learned and its (soft) discriminative output is used as an input for a new (second level) classifier. For example, by doing this we can combine a distance from the hyperplane (from SVM) and tree ratio (from RF) to train a new classifier.

### Evaluation of the classification model quality

The quality of each model using random re-sampling validation schemes [14]. Briefly, our goal is to evaluate the generalization performance of the model, that is, its performance on samples we expect to see in the future. Since these are not available we split the data available to us into the training and test set. The model is always learned on the training set and tested on the test set. We use random sub-sampling approach to divide the dataset using 70/30 train/test split. To avoid potential biases due to a single train/test split we repeat the analysis on multiple (40) random splits and average their results. Each classification model is evaluated using: the accuracy, sensitivity and specificity of the model under 0-1 loss function and the Area under the ROC curve.

## RESULTS AND DISCUSSION

**Table 1** displays classification error (ACE), sensitivity, specificity statistics obtained by the SVM, CART, NB and RF models on MS proteomic data and the Luminex data both individually and after they are (data) merged. The first three statistics are obtained under the 0-1 classification loss model. **Figure 1** and **Figure 2** show ROC curves, their AUC statistics for the RF and SVM models. All reported statistics are averages over 40 independent train/test splits of the entire dataset.

The results show that if the data are taken individually the RF model performs better on the Luminex data, while SVM is better on MS data. The fact that these two models perform differently can be explained by the differences in the measurements these data-sources represent. The MS data are high dimensional and involve a rather high number of correlated signals. On the other hand, the Luminex data are lower dimensional and represent independent probes and their measurements are less correlated. If the data are taken together both models lose their performance. The RF appears to be overwhelmed by the dimensionality of the SELDI data, while linear SVM is limited to linear decision boundary which does not appear to capture well discriminative signals in Luminex data. All these results demonstrate the difficulties arising from the straightforward combination of the two data sources.

| | | Error | SN | SP |
|---|---|---|---|---|
| LUMINEX | CART | 25.66% | 63.31% | 89.47% |
| | std | 18.90% | 31.93% | 22.68% |
| | NB | 23.09% | 60.52% | 94.33% |
| | std | 12.07% | 22.30% | 6.14% |
| | LogisticR | 21.76% | 75.94% | 80.74% |
| | std | 5.64% | 10.99% | 10.95% |
| | RF | **9.41%** | 91.93% | 89.89% |
| | std | 4.77% | 6.56% | 8.79% |
| | SVM | 23.97% | 72.11% | 80.08% |
| | std | 7.37% | 11.33% | 10.08% |
| SELDI PEAKS | CART | 36.99% | 59.49% | 68.32% |
| | std | '8.32% | 17.44% | 20.28% |
| | NB | 45.29% | 83.71% | 25.24% |
| | std | '8.73% | 8.18% | 12.95% |
| | LogisticR | 36.69% | 63.56% | 63.48% |
| | std | 8.48% | 15.10% | 10.88% |
| | RF | 32.35% | 70.32% | 67.23% |
| | std | 9.18% | 14.37% | 16.45% |
| | SVM | **16.62%** | 83.04% | 84.71% |
| | std | 7.37% | 10.88% | 9.77% |
| SELDI PEAKS + LUMINEX | CART | 22.94% | 68.00% | 88.87% |
| | std | 16.07% | 25.71% | 22.71% |
| | NB | 44.63% | 74.21% | 37.74% |
| | std | 9.97% | 26.28% | 25.40% |
| | LogisticR | 38.38% | 60.72% | 62.56% |
| | std | 9.30% | 12.51% | 12.11% |
| | RF | **21.54%** | 76.58% | 82.37% |
| | std | 7.98% | 13.67% | 13.22% |
| | SVM | 34.49% | 50.82% | 79.83% |
| | std | 12.12% | 36.76% | 22.42% |

**Table 1: Data Fusion**

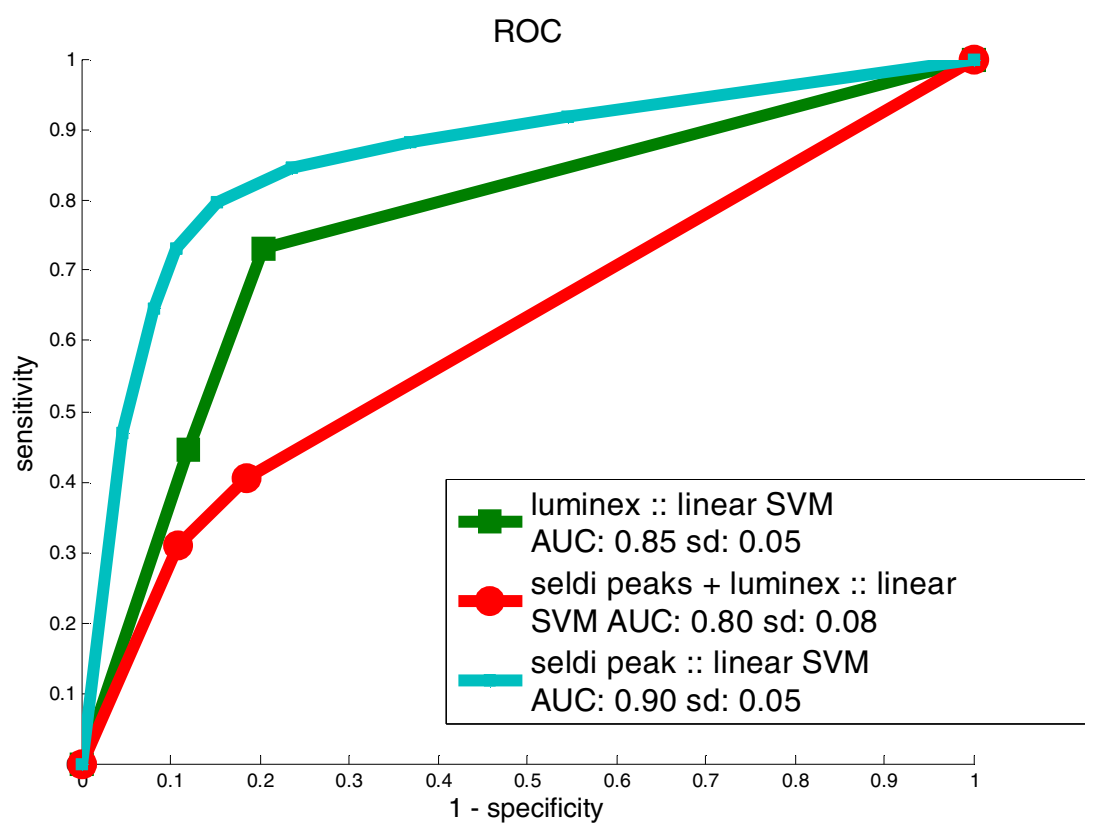

**Figure 1: ROC for linear SVM**

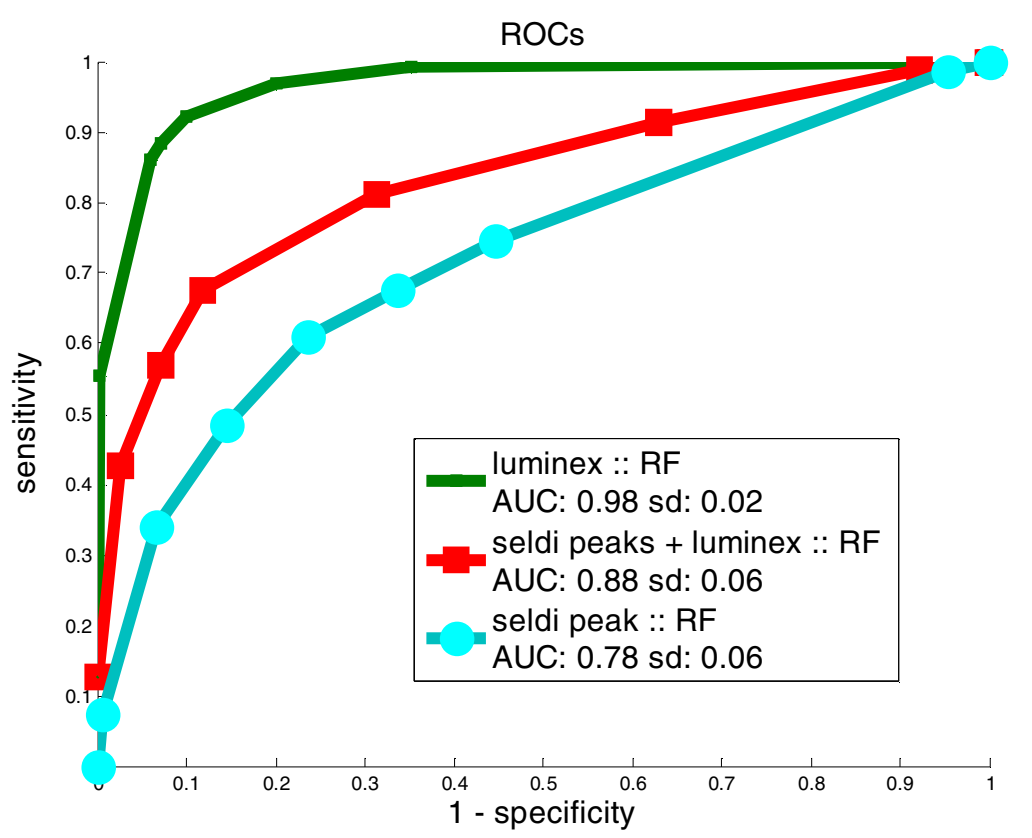

**Figure 2: ROC for Random Forest**

**Table 2** compares different model-combination solutions proposed for the data combination problem. The model composition approach is implemented with the Naïve Bayes model as a second level classifier. For comparison sake we have included also the result for the RF classifier that was trained on Luminex data and top 50 differential features identified in the SELDI-TOF-MS data. The differential feature selection was based on the t-test scoring. Although, the difference between those ROC curves is not sta-tistically significant, each one of them is significantly better than the best ROC for the data fusion. For example, Figure 3. compares the ROCs for the best data fusion and the best model fusion scenario. The ROC for the model fusion model is significantly better ($p < 0.0086$) than the one for the Random Forest on merged data. For the significance analysis, we used variance corrected resampled paired t-test [15].

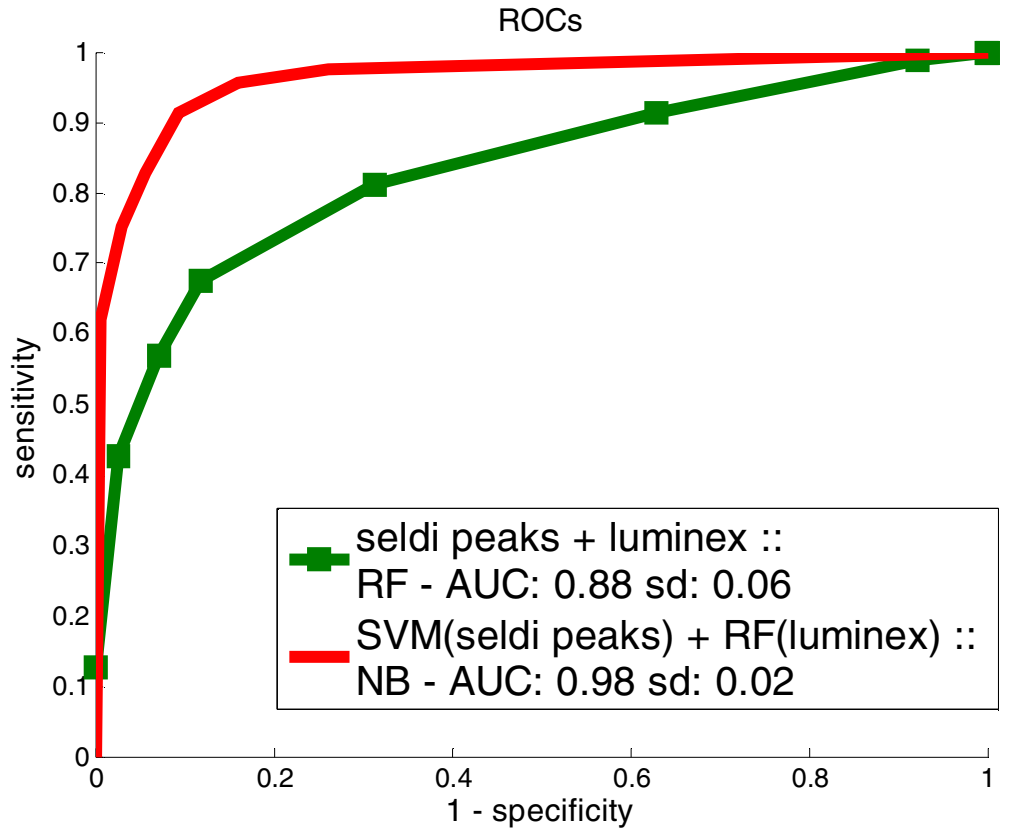

**Figure 3: Comparison of the ROC curves for the best data fusion and the best model fusion model**

| | | Error | SN | SP |
|---|---|---|---|---|
| **SVM(seldi) + RF(luminex)** | NB | 8.82% | 91.28% | 91.60% |
| | std | 4.42% | 7.90% | 7.91% |
| **SVM(seldi) + luminex** | RF | 9.71% | '91.29% | 89.88% |
| | Std | 4.53% | 7.22% | 8.40% |
| **seldi peaks + RF(luminex)** | SVM | 8.46% | 92.56% | 91.03% |
| | Std | 3.78% | 5.98% | 6.58% |
| **T_test50(seldi) + luminex** | RF | 9.85% | 88.83% | 92.12% |
| | std | 4.78% | 9.23% | 7.24% |

**Table 2: Model Combination**

The results in **Table 2** and **Figure 3** demonstrate improvement with the model combination approach over the classification when we merely merged the data. Compound classifiers seem to be able to merge the discriminative information along with merging the data. However, we note that the model combination results are only slightly better or comparable to plain luminex classification. This indicates that most of the discriminative information is offered by the Luminex panel.

## CONCLUSION

With many new bioinformatics data-sources that will emerge in years to come, building of improved disease detection models by utilizing multiple data sources becomes a necessity. In this work we showed that simple data merge may not be the solution to the problem and that it may prevent us from maximizing the benefit of information in the data. More specifically, we showed that the data merge for two proteomic data sources and a fixed classification model may lead to the deterioration of classifier's predictive performance over individual data sources. To address the problem we studied an alternative classification solution in which models built for different data-sources and not data are combined. We illustrated the benefit of the model approach on our two-source dataset. While the results of this initial study are hard to generalize, the mere existence of the problem shows that full utilization of multiple data-sources in classification tasks may go beyond the simple data merge solution .

## Acknowledgments

This work was supported by the DoD grant USAMRAA W81XWH-05-2-0066, NLM training grant 5 T15 LM007059-20 and NCI grant P50 CA090440-06. The authors would like to thank Drs. Bigbee, Zeh, and Whitcomb for the data used in our analyses.